\documentclass[11pt]{article}
\usepackage[final]{acl}
\usepackage{times}
\usepackage{latexsym}
\usepackage[T1]{fontenc}
\usepackage[utf8]{inputenc}
\usepackage{microtype}
\usepackage{inconsolata}
\usepackage{graphicx}
\usepackage{placeins}
\usepackage{times}
\usepackage{latexsym}
\usepackage[T1]{fontenc}
\usepackage[utf8]{inputenc}
\usepackage{microtype}
\usepackage{inconsolata}
\usepackage{graphicx}

\usepackage{xspace,mfirstuc,tabulary}
\usepackage[utf8]{inputenc}
\usepackage{pifont} 
\usepackage{todonotes} 
\usepackage{bbding}
\usepackage{soul} 
\usepackage{xcolor,colortbl} 
\usepackage{appendix} 
\usepackage{enumitem} 
\usepackage{booktabs} 
\usepackage{algorithm} 
\usepackage{multirow} 
\usepackage{array} 
\usepackage{arydshln} 
\usepackage{amsmath, amssymb} \usepackage{algpseudocode} 
\usepackage{hyperref}
\usepackage[table]{xcolor}
\usepackage[normalem]{ulem}
\usepackage[most]{tcolorbox}
\usepackage{fontawesome5}

\newtcolorbox{promptbox}[1]{
    breakable,
    colback=gray!3,
    colframe=black!60,
    sharp corners,
    boxrule=0.4pt,
    left=4pt,
    right=4pt,
    top=4pt,
    bottom=4pt,
    fonttitle=\bfseries,
    title=#1
}

\definecolor{ForestGreen}{rgb}{0.13, 0.55, 0.13}
\definecolor{darkred}{RGB}{139, 0, 0}

\newcommand{\model}{\texttt{TITAN}}

\title{Translation Circuit Discovery and Targeted Optimization for Low-Resource Multilingual Translation}

\author{Atai Waris Khan, Abdullah Mazhar, Raghu Kumar, \textbf{Md Shad Akhtar}\\
  Indraprastha Institute of Information Technology Delhi;\\
  \texttt{\{ataik,abdullahm,raghu22386,shad.akhtar\}@iiitd.ac.in}\\ 
}

\begin{document}
\maketitle
\begin{abstract}
Recent multilingual large language models achieve strong translation performance, yet the internal mechanisms responsible for low-resource multilingual translation remain insufficiently understood. In this paper, we study translation circuits in multilingual decoder-only LLMs for Hindi$\rightarrow$Urdu and Hindi$\rightarrow$Sindhi (Devanagari). We propose \model{}, a mechanistic framework that combines causal intervention, sparse feature decomposition, and targeted preference optimization to identify and optimize translation-crucial transformer components. Unlike conventional full-parameter fine-tuning, \model{} restricts optimization to sparse causally important circuits, enabling scalable and parameter-efficient adaptation while updating fewer than 5\% of model parameters. Experimental results show that \model{} achieves BLEU improvements of +22.67\% on Hindi$\rightarrow$Urdu and +45.86\% on Hindi$\rightarrow$Sindhi over the multilingual baselines. Human expert evaluation further confirms substantial gains in adequacy and fluency across both language pairs. Additional analyses reveal that the identified translation circuits for the studied Indic language pairs operate over a shared semantic subspace rather than through an English-centric pivot representation. Overall, our findings demonstrate that mechanistic circuit discovery can directly improve low-resource multilingual translation while providing interpretable insights into multilingual transformer behavior.
\end{abstract}

\section{Introduction}

Recent multilingual large language models have demonstrated strong translation capabilities across many high-resource language pairs~\cite{zhu2024multilingualmachinetranslationlarge,xu2024paradigmshiftmachinetranslation,alves2024toweropenmultilinguallarge}. At the same time, mechanistic interpretability studies have begun analyzing how translation behavior emerges inside transformer models through circuit analysis and causal intervention~\cite{zhang2026exploringtranslationmechanismlarge}. However, existing work remains largely focused on English-centric and high-resource settings, leaving the mechanisms underlying low-resource multilingual translation insufficiently explored. This limitation is particularly important for South Asian languages, which often exhibit strong lexical overlap, shared grammatical structure, script variation, and phonological divergence~\cite{subbarao2012south}. Consequently, current fine-tuning approaches optimize model parameters without identifying which components are causally responsible for translation behavior.

In this work, we study multilingual translation circuits in decoder-only LLMs for two low-resource South Asian directions: Hindi$\rightarrow$Urdu and Hindi$\rightarrow$Sindhi (Devanagari). These language pairs provide complementary linguistic settings: Hindi and Urdu share substantial lexical and grammatical overlap but differ primarily in script, whereas Hindi and Sindhi additionally differ in phonology and lexical composition due to Sindhi-specific implosive consonants and Persian-Arabic lexical influence~\cite{raza2004phonemic}. This setting enables controlled analysis of whether related low-resource translations rely on shared or specialized computational mechanisms.

To address this problem, we propose \model{} (\textbf{T}ranslation-aware \textbf{I}nterpretability through \textbf{T}argeted \textbf{A}ctivation \textbf{N}etworks), a mechanistic framework for parameter-efficient multilingual translation. \model{} combines CIRCE (\textbf{C}ausal \textbf{I}ntervention for \textbf{R}epresentation \textbf{C}ircuit \textbf{E}ngineering), sparse feature decomposition, and targeted preference optimization to identify and optimize translation-crucial transformer components. Unlike conventional full-parameter fine-tuning, \model{} restricts optimization to sparse causally important circuits, improving both interpretability and parameter efficiency.

\paragraph{\bf Translation Circuit.}
We define a translation circuit as a sparse set of transformer components that causally contribute to a specific translation behavior. These components are identified through causal intervention by measuring their effect on translation-relevant outputs, yielding a circuit comprising causally important attention heads and MLP blocks for the studied translation task.

Our analysis reveals that the identified translation circuits for the studied low-resource Indic language pairs are distributed across attention and MLP components. We further show that the identified translation circuits for the studied Indic translation directions operate through a shared semantic subspace rather than through the English-centric pivot mechanisms reported in prior work~\cite{zhang2026exploringtranslationmechanismlarge}. Experimentally, \model{} achieves BLEU improvements of +33.75\% on Hindi$\rightarrow$Urdu and +45.86\% on Hindi$\rightarrow$Sindhi while updating fewer than 5\% of model parameters. Human expert evaluation further confirms substantial improvements in adequacy and fluency across both language pairs. We additionally validate the generalisability of the discovered sparse-circuit behavior on English$\leftrightarrow$French and English$\leftrightarrow$Hindi translation directions.

\paragraph{Contributions.} Our contributions are summarized as follows:
\begin{itemize}[leftmargin=*]

    \item We propose \model{}, a mechanistic framework that combines causal intervention, sparse feature decomposition, and targeted preference optimization for parameter-efficient multilingual translation.

    \item We show that the identified translation circuits for the studied low-resource Indic language pairs operate through a shared semantic subspace rather than an English-centric pivot mechanism.

    \item We demonstrate that restricting optimization to causally identified translation circuits substantially improves translation quality while updating fewer than 5\% of model parameters.
\end{itemize}

\paragraph{Reproducibility.} Code and dataset is open-sourced on \href{https://anonymous.4open.science/r/TITAN-9BEA/README.md}{
  \faGithub\ GitHub \faExternalLink*\ }.
\section{Related Work}

Low-resource neural machine translation, mechanistic interpretability, and multilingual large language models have emerged as closely connected research directions. Prior work has improved multilingual transfer and translation quality for underrepresented languages, while recent mechanistic interpretability research enables causal analysis of transformer computations through activation intervention and sparse feature decomposition.

\paragraph{\bf Low-Resource Neural Machine Translation.}
Low-resource neural machine translation (NMT) remains challenging due to limited parallel corpora and linguistic diversity across Indic languages. Recent work has explored multilingual transfer learning, back-translation, transliteration, and pivot-based modeling to improve translation quality for low-resource Indian languages~\cite{wei-etal-2024-machine,p-m-etal-2024-mtnlp,10.1145/3652026}. Recent parameter-efficient approaches further investigate language-specific adaptation, including language-specific LoRA, intrinsic language-specific subspaces, and language-aware parameter selection for multilingual knowledge transfer. These methods improve language-specific adaptation but primarily focus on parameter specialization rather than identifying components that are causally responsible for translation behavior.

\paragraph{\bf Mechanistic Interpretability and Multilingual LLMs.}
Mechanistic interpretability seeks to reverse-engineer transformer computations through causal interventions and circuit analysis. Prior work has used activation and path patching to identify circuits underlying reasoning and attention. In multilingual neural machine translation, studies have identified language-specific subspaces emerging during multilingual fine-tuning \cite{cao2024exploring}, as well as language-aware parameters for targeted cross-lingual knowledge transfer \cite{dong2025mlas}. Language-specific fine-tuning has further been explored for adapting LLMs to low-resource translation settings \cite{liang2025toward}. Recent circuit-discovery work, such as Emotion Circuits \cite{wang2025llms}, combines contrastive examples with causal analysis to identify functionally relevant representations. Unlike these approaches, \model{} focuses on discovering translation-specific circuits and leveraging them for targeted multilingual translation optimization.

\begin{table*}[t]
\centering
\small

\setlength{\tabcolsep}{5pt}
\renewcommand{\arraystretch}{1.15}

\begin{tabular}{llrrrrrr}
\toprule

\multirow{2}{*}{\textbf{Pair}} &
\multirow{2}{*}{\textbf{Split}} &
\multirow{2}{*}{\textbf{\#Sent}} &
\multicolumn{2}{c}{\textbf{Hindi}} &
\multicolumn{2}{c}{\textbf{Target Language}} &
\multirow{2}{*}{\textbf{Avg.\ Len.}} \\

\cmidrule(lr){4-5}
\cmidrule(lr){6-7}

&
&
&
\textbf{\#Tokens} &
\textbf{\#Types} &
\textbf{\#Tokens} &
\textbf{\#Types} &
\\

\midrule

\multirow{3}{*}{\textbf{Hi--Sd}}

& SFT Train
& 20K
& 404.6k
& 36.1K
& 398.2K
& 49.7K
& 20.1 \\

& DPO Train
& 15K
& 302.8K
& 30.5K
& 297.8K
& 41.6K
& 20.0 \\

& Contrastive-From SFT
& 10K
& 220.6K
& 15.8K
& 217.6K
& 20.5K
& 21.9 \\

\midrule

\multirow{3}{*}{\textbf{Hi--Ur}}

& SFT Train
& 30K
& 562.3K
& 47.3K
& 599.2K
& 43.7K
& 19.4 \\

& DPO Train
& 30K
& 567.1K
& 47.4K
& 603.1K
& 43.8K
& 19.5 \\

& Contrastive-From SFT
& 10K
& 186.9K
& 15.6K
& 198.9K
& 15.1K
& 19.3 \\

\bottomrule
\end{tabular}
\caption{
Corpus statistics for the Hindi--Sindhi (Devanagari) and Hindi--Urdu parallel corpora used in TITAN. Each corpus is partitioned into SFT, DPO, and contrastive analysis splits. We sample a 10K contrastive subset from the SFT split of each language
pair to construct contrastive-based intervention prompts for causal translation circuit identification. \textbf{Hi--Sd}: Hindi--Sindhi (Devanagari);
\textbf{Hi--Ur}: Hindi--Urdu.
Avg.\ Len.\ denotes the mean number of whitespace-tokenized tokens per sentence.
}
\label{tab:corpus_stats}

\end{table*}

\paragraph{\bf Preference Optimization for Translation Alignment.}
Preference optimization methods such as Direct Preference Optimization have recently been used to improve multilingual generation and translation quality through pairwise preference learning. Existing approaches generally optimize model parameters without identifying which components are causally responsible for translation behavior. In contrast, \model{} combines mechanistic circuit discovery with targeted optimization to selectively optimize translation-relevant components.

\section{Dataset}
\label{sec:data}

We evaluate \model{} on two low-resource South Asian translation pairs: Hindi--Urdu and Hindi--Sindhi (Devanagari). These language pairs provide complementary multilingual settings involving script divergence, lexical overlap, and phonological variation. Corpus statistics are summarized in Table~\ref{tab:corpus_stats}.

\subsection{Translation Corpora}
\label{ssec:data}

We use the Hindi--Urdu and Hindi--Sindhi subsets of the Bharat Parallel Corpus Collection (BPCC)~\cite{gala2023indictrans}. Hindi and Urdu share substantial grammatical and lexical overlap but differ primarily in script, with Hindi written in Devanagari and Urdu in Arabic-Nastaliq. In contrast, although Hindi and Sindhi both use the Devanagari script, Sindhi additionally contains implosive consonants and stronger Persian-Arabic lexical influence.

After preprocessing and deduplication, the Hindi--Urdu corpus contains 60K parallel sentence pairs, while the Hindi--Sindhi corpus contains 35K sentence pairs. Both corpora are partitioned into supervised fine-tuning (SFT), Direct Preference Optimization (DPO), and contrastive analysis splits.

\subsection{Contrastive Translation Data}
\label{ssec:contrastive}

To support mechanistic analysis, we construct contrastive translation pairs consisting of valid translation prompts and counterfactual prompts where translation intent is removed while preserving grammatical structure. We sample 10K sentence pairs from the SFT split of each language pair and generate counterfactual variants using semantic perturbation strategies. For Hindi--Sindhi, sampling is additionally stratified to ensure sufficient coverage of Sindhi-specific implosive phoneme markers. Detailed prompt templates, perturbation taxonomies, and validation protocols are provided in Appendix Section~\ref{app:construction}.

\subsection{Preprocessing}
\label{ssec:preprocessing}

We apply uniform preprocessing across both corpora, including Unicode normalization, duplicate removal, and sentence-length filtering. Hindi and Sindhi samples are validated using Devanagari Unicode ranges, whereas Urdu samples are verified using Arabic-script code points. For Hindi--Sindhi, we additionally preserve Sindhi-specific implosive phoneme markers to support phonological circuit analysis.

\section{Methodology}
\label{sec:method}

We propose \textbf{\model{}} (\textbf{T}ranslation-aware \textbf{I}nterpretability through \textbf{T}argeted \textbf{A}ctivation \textbf{N}etworks), a mechanistic interpretability-guided framework for parameter-efficient multilingual translation. Rather than updating all transformer parameters, \model{} identifies the sparse set of components causally responsible for translation behavior and restricts optimization to the discovered translation circuits. The framework is built upon CIRCE (\textbf{C}ausal \textbf{I}ntervention for \textbf{R}epresentation \textbf{C}ircuit \textbf{E}ngineering), which combines subspace intervention with sparse feature decomposition to isolate translation-relevant computation. All CIRCE and \model{} experiments use Gemma-2 9B as the underlying base model.

Hindi--Urdu and Hindi--Sindhi provide complementary multilingual settings for circuit analysis. Hindi--Urdu exhibits strong grammatical overlap with script divergence, whereas Hindi--Sindhi shares script similarity but differs substantially in phonology and lexical composition due to Sindhi-specific implosive phonemes and Persian-Arabic lexical influence. This setting enables \model{} to analyze how translation circuits specialize across distinct forms of linguistic variation.

\subsection{Contrastive Translation Construction}
\label{sec:contrastive}

To isolate translation-specific computation from general language modeling behavior, we construct contrastive translation pairs consisting of a valid translation prompt $X^{+}$ and a counterfactual prompt $X^{-}$:
\begin{equation}
X^{+} = \texttt{``[SRC]: [text] - [TGT]:''}
\end{equation}

Counterfactual prompts preserve grammatical structure while removing translation intent through target nullification, semantic distortion, and contradictory task descriptors. This design allows \model{} to separate activations associated with translation generation from those associated with general multilingual processing.

Each sample additionally contains translation direction labels, vocabulary overlap annotations, and phoneme-level markers. For Hindi--Sindhi, we explicitly annotate samples containing Sindhi-specific implosive phonemes to support phonological circuit analysis.

\subsection{CIRCE: Translation Circuit Identification}
\label{sec:circe}

\model{} identifies translation-relevant transformer components using CIRCE, a subspace-constrained extension of activation patching. For a transformer component $c$ at layer $\ell$, we compute activation differences between positive and counterfactual prompts:

\begin{equation}
\Delta \mathbf{a}_{c}^{(i)}
=
\mathbf{a}_{c}(X_i^{+})
-
\mathbf{a}_{c}(X_i^{-})
\end{equation}

The resulting activation differences are aggregated into a component-specific activation matrix $\mathbf{M}_{c}$ and decomposed into translation-relevant structure and residual variation:

\begin{equation}
\small
\begin{aligned}
\min_{\mathbf{S}_c,\mathbf{E}_c,\mathbf{\Gamma}}
\quad
&
\left\|
\mathbf{M}_c
-
\mathbf{S}_c\mathbf{1}^{\top}
-
\mathbf{E}_c\mathbf{\Gamma}^{\top}
\right\|_F
\\
\text{s.t.}
\quad
&
\text{Span}(\mathbf{S}_c)
\perp
\text{Span}(\mathbf{E}_c)
\end{aligned}
\end{equation}

Here, $\mathbf{S}_{c}$ denotes a single translation-relevant direction, while $\mathbf{E}_{c}$ captures residual variation; this direction defines the intervention basis $\mathbf{U}_{c}$.
Instead of replacing the full activation, CIRCE intervenes only along the translation-relevant direction:

\begin{small}
\begin{align}
\tilde{\mathbf{a}}_c
=
\mathbf{U}_c \mathbf{U}_c^\top
\mathbf{a}_c(X^-)
+
(\mathbf{I}-\mathbf{U}_c \mathbf{U}_c^\top)
\mathbf{a}_c(X^+)
\end{align}
\end{small}

Component importance is measured using intervention-induced translation logit changes. Rather than using fixed thresholds, \model{} calibrates component selection from the empirical intervention distribution using a P99 noise-floor criterion, yielding an operating threshold near $0.25\%$. This preserves sparse but reproducible low-resource translation signals while filtering negligible intervention effects. Components producing substantial translation degradation under intervention are retained as translation-crucial, forming the translation circuit used for optimization; CIRCE applies targeted SFT to these components while freezing all others.

For interpretability, discovered attention heads are further categorized into source, indicator, and positional heads based on their dominant routing behavior.

\subsection{Sparse Feature Decomposition}
\label{sec:sae}

Transformer activations frequently exhibit feature superposition, where multiple linguistic properties are encoded within the same representation. To disentangle these representations, \model{} trains Sparse Autoencoders (SAEs) exclusively on residual activations extracted from CIRCE-identified translation-crucial layers. Restricting SAE training to CIRCE-identified layers ensures that the discovered sparse features remain grounded in causally relevant translation circuitry rather than generic multilingual representations.

For an activation vector $\mathbf{h}^{(\ell)}$, the encoder computes a sparse latent representation:

\begin{equation}
\mathbf{z}^{(\ell)}
=
\text{ReLU}
\left(
(\mathbf{W}_{\ell})^{\top}\mathbf{h}^{(\ell)}
+
\mathbf{b}_{\ell}
\right)
\end{equation}

The original activation is reconstructed through:

\begin{equation}
\hat{\mathbf{h}}^{(\ell)}
=
\mathbf{W}_{\ell}\mathbf{z}^{(\ell)}
\end{equation}

Training minimizes a joint reconstruction and sparsity objective:

\begin{equation}
\mathcal{L}_{\text{SAE}}
=
\|
\mathbf{h}^{(\ell)}
-
\hat{\mathbf{h}}^{(\ell)}
\|_2^2
+
\lambda
\|
\mathbf{z}^{(\ell)}
\|_1
\end{equation}

The resulting sparse representations expose disentangled translation features that are substantially easier to interpret than raw transformer activations.

\subsection{Feature-Level Causal Scoring}
\label{sec:feature}

We evaluate each SAE feature $z_j$ using language selectivity and causal lift. To isolate language identity from semantic variation, selectivity is computed using semantically matched bilingual prompts expressing the same meaning in different languages:

\begin{equation}
\widetilde{\text{Sel}}_j
=
\frac{
\mathbb{E}[z_j(x_{\text{tgt}})]
-
\mathbb{E}[z_j(x_{\text{src}})]
}{
\text{Std}[z_j(x_{\text{tgt}})]
+
\text{Std}[z_j(x_{\text{src}})]
+
\epsilon
}
\end{equation}

To estimate causal relevance beyond correlation, we intervene on feature $j$ during decoding and measure the induced translation logit change:

\begin{equation}
\text{Lift}_j^{(\ell)}
=
\mathbb{E}_{x \sim D}
\left[
\frac{1}{|T|}
\sum_{t \in T}
\Delta M_{\alpha}^{(\ell,j)}(x,t)
\right]
\end{equation}

where $T$ denotes decoding steps and $\Delta M_{\alpha}^{(\ell,j)}$ represents the translation logit change induced by intervention magnitude $\alpha$.

We compute a joint feature score:
\begin{align}
\text{Score}_j = \max(\widetilde{\text{Sel}}_j,0) \cdot \max(\text{Lift}_j,0)
\end{align}

Features exhibiting both high selectivity and positive causal lift are retained as translation-relevant features, while low-scoring features are discarded. This filtering stage provides a finer-grained characterization of the CIRCE-identified translation circuit by distinguishing causally relevant features from language-selective but causally weak features.

\subsection{Targeted Translation Optimization}
\label{sec:optimization}

Using the discovered translation circuit, CIRCE performs targeted SFT on the identified components while freezing all others, whereas \model{} further applies DPO to the same components. Training proceeds hierarchically from word-level translation to phrase-level and finally sentence-level translation, allowing the model to gradually acquire increasingly complex translation structure.

To preserve the general language capabilities of the base model, optimization is regularized using a KL divergence constraint:

\begin{equation}
\mathcal{L}_{\text{total}}
=
\mathcal{L}_{\text{MT}}
+
\lambda_{\text{KL}}
\text{KL}(p_{\theta} \| p_{\text{base}})
\end{equation}

where $\mathcal{L}_{\text{MT}}$ denotes the standard translation objective.

After supervised fine-tuning, \model{} further refines translation quality using Direct Preference Optimization. Preference pairs are constructed by treating the gold reference translations from the parallel corpus as preferred outputs, while translations generated by the supervised fine-tuned model are treated as rejected outputs. 

\section{Experiments and Results}

Here, we compare \model{} against multilingual machine translation systems and decoder-only large language models across Hindi-Urdu and Hindi-Sindhi (Devanagari) translation tasks. We further analyze the effectiveness of targeted circuit optimization under both SFT and DPO settings. In addition to quantitative evaluation, we provide mechanistic analysis to study the role of sparse translation circuits in low-resource multilingual translation.

\paragraph{Baselines.}
We compare \model{} against both multilingual neural machine translation systems and instruction-tuned decoder-only LLMs under a unified evaluation setup. The evaluated baselines include NLLB-200 and IndicTrans2, which represent strong multilingual translation models specifically designed for low-resource settings, as well as Gemma-2, Qwen-3.5, LLaMA-3.1, and Mistral3, which represent recent decoder-only multilingual LLMs. All models are evaluated under zero-shot, supervised fine-tuning (SFT), and DPO settings. CIRCE denotes targeted SFT on the causally identified translation-crucial components, while \model{} further applies DPO to the same components.

\begin{table*}[t]
\centering
\footnotesize
\setlength{\tabcolsep}{2.8pt}
\renewcommand{\arraystretch}{1.08}

\resizebox{\textwidth}{!}{

\begin{tabular}{
>{\centering\arraybackslash}p{0.42cm}
l
*{16}{>{\centering\arraybackslash}p{0.72cm}}
}

\toprule

&
&
\multicolumn{8}{c}{\cellcolor{gray!10}\textbf{Hindi--Urdu}}
&
\multicolumn{8}{c}{\cellcolor{gray!10}\textbf{Hindi--Sindhi (Devanagari)}}
\\

\cmidrule(lr){3-10}
\cmidrule(lr){11-18}

&
&
\multicolumn{4}{c}{\textbf{H$\rightarrow$U}}
&
\multicolumn{4}{c}{\textbf{U$\rightarrow$H}}
&
\multicolumn{4}{c}{\textbf{H$\rightarrow$S}}
&
\multicolumn{4}{c}{\textbf{S$\rightarrow$H}}
\\

\cmidrule(lr){3-6}
\cmidrule(lr){7-10}
\cmidrule(lr){11-14}
\cmidrule(lr){15-18}

&
\textbf{Models}

&
\textbf{BL}
&
\textbf{CF2}
&
\textbf{TER}
&
\textbf{COM}

&
\textbf{BL}
&
\textbf{CF2}
&
\textbf{TER}
&
\textbf{COM}

&
\textbf{BL}
&
\textbf{CF2}
&
\textbf{TER}
&
\textbf{COM}

&
\textbf{BL}
&
\textbf{CF2}
&
\textbf{TER}
&
\textbf{COM}
\\

\midrule

\multirow{6}{*}{\rotatebox[origin=c]{90}{\textbf{ZS}}}
& NLLB-200
& 0.01 & 0.09 & 105.65 & 57.08
& 0.31 & 0.61 & 108.63 & 55.39
& 0.44 & 1.05 & 101.23 & 41.34
& 2.31 & 10.58 & 91.36 & 43.08
\\

& IndicTrans2
& 4.33 & 25.19 & 80.69 & 62.85
& 2.12 & 23.63 & 87.16 & 60.57
& 0.59 & 3.08  & 119.88& 41.76
& 1.15 & 7.38 & 109.90 & 55.16
\\

& Mistral-3
& 0.32 & 2.40 & 103.46 & 33.70
& 0.19 & 2.56 & 102.95 & 30.12
& 0.17 & 1.92 & 296.8 & 33.43
& 0.68 & 5.21 & 238.11 & 27.63
\\

& LLaMA-3.1
& 1.46 & 3.78 & 117.57 & 43.83
& 0.11 & 4.20 & 100.79 & 44.42
& 0.51 & 3.63 & 273.22 & 37.52
& 1.21 & 6.93 & 250.01 & 33.43
\\

& Qwen-3.5
& 0.21 & 3.02 & 99.23 & 40.30
& 0.53 & 3.30 & 99.02 & 41.22
& 0.21 & 3.02 & 99.23 & 34.80
& 0.53 & 3.30 & 109.02 & 31.20
\\

& Gemma-2
& 5.93 & 28.23 & 69.81 & 59.05
& 6.40 & 25.58 & 79.30 & 53.29
& 0.62 & 5.93 & 187.74 & 39.87
& 2.26 & 8.42 & 79.30 & 36.41
\\

\midrule

\multirow{6}{*}{\rotatebox[origin=c]{90}{\textbf{SFT}}}
& NLLB-200
& 23.39 & 56.61 & 41.70 & 72.76
& 22.03 & 51.20 & 44.77 & 73.73
& 15.89 & 33.42 & 88.22 & 56.83
& 17.33 & 31.29 & 78.05 & 65.89
\\

& IndicTrans2
& 23.63 & 41.43 & 76.24 & 70.87
& 25.36 & 50.33 & 49.36 & 74.38
& 14.83 & 34.57 & 78.65 & 60.2
& 18.66 & 48.96 & 81.23 & 67.14
\\

& LLaMA-3.1
& 10.59 & 36.58 & 59.87 & 66.8
& 9.59 & 31.32 & 70.21 & 65.35
& 9.81 & 23.22 & 95.99 & 58.98
& 10.25 & 18.59 & 96.76 & 57.38
\\

& Qwen-3.5
& 8.89 & 33.90 & 73.27 & 64.05
& 7.25 & 30.86 & 73.15 & 51.63
& 5.63 & 17.10 & 110.99 & 38.52
& 7.21 & 21.85 & 99.02 & 45.31
\\

& Gemma-2
& 23.79 & 56.66 & 40.98 & 74.16
& 25.89 & 54.69 & 43.75 & 75.33
& 15.94 & 34.10 & 86.50 & 59.86
& 17.38 & 47.56 & 75.28 & 72.77
\\

& Mistral-3
& 9.47 & 31.94 & 72.57 & 69.16
& 12.00 & 36.78 & 69.28 & 67.22
& 5.44 & 20.38 & 102.90 & 55.77
& 5.58 & 23.63 & 95.25 & 51.93
\\

\midrule

\multirow{6}{*}{\rotatebox[origin=c]{90}{\textbf{DPO}}}
& NLLB-200
& 25.86 & 59.36 & 40.14 & 74.98
& 23.51 & 55.52 & 43.00 & 75.15
& 17.00 & 40.92 & 58.22 & 58.39
& 18.34 & 50.15 & 51.13 & 70.64
\\

& IndicTrans2
& 24.68 & 58.25 & 50.22 & 71.52
& 24.96 & 56.30 & 40.58 & 74.84
& 17.32 & 40.30 & 70.60 & 61.9
& 20.63 & 50.91 & 72.55 & 68.56
\\

& LLaMA-3.1
& 11.29 & 39.18 & 53.63 & 68.37
& 9.93 & 34.50 & 57.70 & 65.83
& 10.21 & 27.66 & 72.36 & 59.89
& 12.11 & 23.32 & 70.39 & 61.85
\\

& Qwen-3.5
& 8.90 & 33.70 & 71.64 & 64.98
& 7.32 & 30.90 & 71.21 & 52.74
& 8.98 & 29.31 & 96.44 & 44.96
& 8.36 & 30.02 & 87.37 & 50.04
\\

& Gemma-2
& 25.94 & 60.10 & 42.37 & 75.88
& 25.78 & 56.10 & 44.39 & 77.37
& 18.01 & 43.68 & 84.71 & 62.81
& 19.89 & 51.20 & 69.25 & 74.55
\\

& Mistral-3
& 9.29 & 30.61 & 73.60 & 68.84
& 11.22 & 36.10 & 68.28 & 70.01
& 6.22 & 25.58 & 97.50 & 56.88
& 7.76 & 32.84 & 88.25 & 54.69
\\

\midrule

\rowcolor{green!5}

&
\textbf{CIRCE}
&
\textbf{28.03}
&
\textbf{62.47}
&
\textbf{58.56}
&
\textbf{79.33}
&
\textbf{26.23}
&
\textbf{53.88}
&
\textbf{71.36}
&
\textbf{80.52}
&
\textbf{22.19}
&
\textbf{52.90}
&
\textbf{45.21}
&
\textbf{65.76}
&
\textbf{25.47}
&
\textbf{61.48}
&
\textbf{28.36}
&
\textbf{76.5}
\\

\midrule

\rowcolor{green!28}

&
\textbf{TITAN}
&
\textbf{31.82}
&
\textbf{65.93}
&
\textbf{52.30}
&
\textbf{80.52}
&
\textbf{27.01}
&
\textbf{55.44}
&
\textbf{72.10}
&
\textbf{81.28}
&
\textbf{23.25}
&
\textbf{55.92}
&
\textbf{42.52}
&
\textbf{70.17}
&
\textbf{27.80}
&
\textbf{65.33}
&
\textbf{25.54}
&
\textbf{78.95}
\\

\midrule

&
$\Delta_{\text{SFT}}(\%)$
&
\textcolor{blue}{$\uparrow 33.75$}
&
\textcolor{blue}{$\uparrow 5.77$}
&
\textcolor{red}{$\downarrow 27.62$}
&
\textcolor{blue}{$\uparrow 6.11$}

&
\textcolor{blue}{$\uparrow 4.33$}
&
\textcolor{blue}{$\uparrow 1.37$}
&
\textcolor{red}{$\downarrow 60.23$}
&
\textcolor{blue}{$\uparrow 5.05$}

&
\textcolor{blue}{$\uparrow 45.86$}
&
\textcolor{blue}{$\uparrow 61.76$}
&
\textcolor{blue}{$\uparrow 45.94$}
&
\textcolor{blue}{$\uparrow 11.72$}

&
\textcolor{blue}{$\uparrow 48.98$}
&
\textcolor{blue}{$\uparrow 33.44$}
&
\textcolor{blue}{$\uparrow 66.07$}
&
\textcolor{blue}{$\uparrow 5.90$}
\\
\bottomrule

\end{tabular}

}

\caption{
Translation performance comparison across Hindi-Urdu and Hindi-Sindhi (Devanagari) under zero-shot (ZS), supervised fine-tuning (SFT), and Direct Preference Optimization (DPO) settings. TITAN consistently outperforms multilingual translation baselines across both language pairs, demonstrating the effectiveness of sparse causal translation circuits for low-resource multilingual translation. All CIRCE and TITAN results use Gemma-2 9B as the underlying base model. \textbf{BL}: BLEU, \textbf{CF2}: ChrF2, \textbf{TER}: Translation Edit Rate, \textbf{COM}: COMET.
}
\label{tab:joint_results}
\end{table*}
\paragraph{Performance Comparison.}
Table~\ref{tab:joint_results} reports translation performance across Hindi--Urdu and Hindi--Sindhi translation directions. Zero-shot performance remains consistently weak across both multilingual MT systems and decoder-only LLMs, particularly for Hindi--Sindhi due to additional phonological and lexical divergence. Supervised fine-tuning and DPO improve translation quality, with Gemma-2 and NLLB-200 emerging as the strongest multilingual baselines across most directions.

\model{} consistently outperforms all baselines across both language pairs. The gains are most pronounced for Hindi--Sindhi, where \model{} achieves relative BLEU improvements of +45.86\% and +48.98\% for H$\rightarrow$S and S$\rightarrow$H respectively over the strongest baselines. Similar improvements are observed for Hindi--Urdu, demonstrating that restricting optimization to causally identified translation circuits substantially improves low-resource multilingual translation. Notably, the CIRCE-only variant already surpasses several fully optimized baselines despite updating only sparse translation-relevant components. Overall, the results show that \model{} provides more scalable and parameter-efficient alternative to conventional full-model optimization.

For a parameter-controlled comparison, we match LoRA SFT+DPO to \model{} at the same 3.14\% trainable-parameter budget. Despite the identical budget, LoRA SFT+DPO achieves 26.37 BLEU, compared with 31.82 BLEU for \model{} (+5.45 BLEU), indicating that the gain comes from circuit-based component selection rather than parameter count.

\paragraph{\bf Generalizability.}
We evaluate \model{} beyond the low-resource Hindi--Urdu and Hindi--Sindhi settings in two complementary ways. For each evaluation direction, the translation circuit is independently identified using CIRCE on that language pair before targeted optimization. We evaluate \model{} beyond the low-resource Hindi--Urdu and Hindi--Sindhi settings in two complementary ways.

First, we assess high-resource cross-lingual generalization on Hindi--English and English--French. Table~\ref{tab:generalization} reports the results. For Hindi$\rightarrow$English, \model{} improves BLEU from 29.33 with DPO to 34.21, while English$\rightarrow$Hindi improves from 28.71 to 31.94. On English$\rightarrow$French, BLEU increases from 30.58 to 33.57, and French$\rightarrow$English from 33.09 to 36.40. Similar improvements are observed in ChrF++ and TER, indicating consistent gains across evaluation metrics.

\begin{table}[t]
\centering
\footnotesize
\setlength{\tabcolsep}{3.5pt}
\renewcommand{\arraystretch}{1.05}

\begin{tabular}{llcccc}

\toprule

\textbf{Setting}
&
\textbf{Direction}
&
\textbf{BLEU}
&
\textbf{chrF++}
&
\textbf{TER}
&
\textbf{COMET}

\\

\midrule

\multicolumn{5}{c}{\textbf{Hindi--English}}
\\

\midrule

SFT
& Hi$\rightarrow$En
& 27.96
& 52.36
& 70.24
& 73.71
\\

DPO
& Hi$\rightarrow$En
& 29.33
& 54.37
& 66.35
& 75.43
\\

CIRCE
& Hi$\rightarrow$En
& 33.48
& 58.11
& 66.34
& 76.39
\\

\rowcolor{green!18}

TITAN
& Hi$\rightarrow$En
& \textbf{34.21}
& \textbf{60.04}
& \textbf{60.22}
& \textbf{76.72}
\\

\midrule

SFT
& En$\rightarrow$Hi
& 26.08
& 50.88
& 70.49
& 78.15
\\

DPO
& En$\rightarrow$Hi
& 28.71
& 52.34
& 65.91
& 79.6
\\

CIRCE
& En$\rightarrow$Hi
& 31.03
& 56.99
& 65.16
& 82.36
\\

\rowcolor{green!18}

TITAN
& En$\rightarrow$Hi
& \textbf{31.94}
& \textbf{57.51}
& \textbf{62.21}
& \textbf{83.95}
\\

\midrule

\multicolumn{5}{c}{\textbf{English--French}}
\\

\midrule

SFT
& En$\rightarrow$Fr
& 29.64
& 52.88
& 60.44
& 80.01
\\

DPO
& En$\rightarrow$Fr
& 30.58
& 55.24
& 58.63
& 80.88
\\
CIRCE
& En$\rightarrow$Fr
& 32.43
& 55.40
& 59.55
& 84.20
\\

\rowcolor{green!18}

TITAN
& En$\rightarrow$Fr
& \textbf{33.57}
& \textbf{57.90}
& \textbf{54.35}
& \textbf{85.51}
\\

\midrule

SFT
& Fr$\rightarrow$En
& 31.02
& 54.89
& 54.17
& 82.5
\\

DPO
& Fr$\rightarrow$En
& 33.09
& 56.51
& 49.63
& 83.41
\\

CIRCE
& Fr$\rightarrow$En
& 34.87
& 55.72
& 50.13
& 85.98
\\

\rowcolor{green!18}

TITAN
& Fr$\rightarrow$En
& \textbf{36.40}
& \textbf{59.25}
& \textbf{46.08}
& \textbf{86.77}
\\

\bottomrule

\end{tabular}
\caption{
Cross-lingual generalization on unseen high-resource Hindi--English and English--French translation tasks. TITAN consistently outperforms SFT and DPO baselines, using Gemma-2 9B as the base model; CIRCE applies targeted SFT, while TITAN further applies DPO to the same identified components.
}
\label{tab:generalization}
\end{table}
Second, to test whether the observed behavior extends beyond closely related Indic languages, we evaluate \model{} on two typologically distinct low-resource directions: Hindi$\rightarrow$Tamil and English$\rightarrow$Irish. Table~\ref{tab:diverse_lowresource} reports the results. \model{} improves BLEU from 23.01 to 26.71 on Hindi$\rightarrow$Tamil and from 29.52 to 33.19 on English$\rightarrow$Irish over the DPO baseline, corresponding to relative improvements of 16.08\% and 12.43\%, respectively. The consistent performance ordering of SFT $<$ DPO $<$ CIRCE $<$ \model{} across language pairs further supports the generality of circuit-based optimization beyond the closely related Indic language pairs used for circuit discovery.

\begin{table}[t]
\centering
\footnotesize
\setlength{\tabcolsep}{3.5pt}
\renewcommand{\arraystretch}{1.05}

\begin{tabular}{llcccc}
\toprule
\textbf{Setting}
&
\textbf{Direction}
&
\textbf{BLEU}
&
\textbf{chrF++}
&
\textbf{TER}
&
\textbf{COMET}
\\
\midrule

\multicolumn{6}{c}{\textbf{Hindi--Tamil}}
\\
\midrule

SFT
& Hi$\rightarrow$Ta
& 22.61
& 40.66
& 81.66
& 77.20
\\

DPO
& Hi$\rightarrow$Ta
& 23.01
& 42.87
& 77.20
& 79.00
\\

CIRCE
& Hi$\rightarrow$Ta
& 25.04
& 49.46
& 75.80
& 81.20
\\

\rowcolor{green!18}
\textbf{TITAN}
& Hi$\rightarrow$Ta
& \textbf{26.71}
& \textbf{50.60}
& \textbf{72.20}
& \textbf{82.70}
\\

\midrule

\multicolumn{6}{c}{\textbf{English--Irish}}
\\
\midrule

SFT
& En$\rightarrow$Ga
& 26.39
& 51.71
& 84.35
& 70.60
\\

DPO
& En$\rightarrow$Ga
& 29.52
& 53.71
& 80.26
& 72.80
\\

CIRCE
& En$\rightarrow$Ga
& 30.84
& 54.42
& 76.90
& 74.60
\\

\rowcolor{green!18}
\textbf{TITAN}
& En$\rightarrow$Ga
& \textbf{33.19}
& \textbf{57.82}
& \textbf{66.70}
& \textbf{75.70}
\\

\bottomrule
\end{tabular}

\caption{
Performance on additional typologically diverse low-resource translation tasks. TITAN consistently outperforms SFT, DPO, and CIRCE on Hindi--Tamil and English--Irish, demonstrating that the observed gains extend beyond the closely related Indic language pairs used for primary evaluation.
}
\label{tab:diverse_lowresource}
\end{table}
\section{Behavioral Pattern Analysis}

We now investigate the internal behavioral patterns learned by multilingual decoder-only LLMs during low-resource translation. Specifically, we analyze whether \model{} identifies genuinely causal translation circuits, how translation responsibilities are distributed across transformer components, and how multilingual representations evolve across layers. Unless otherwise stated, analyses are performed on the strongest \model{} checkpoint for both Hindi$\rightarrow$Urdu and Hindi$\rightarrow$Sindhi (Devanagari).
\subsection{Causal Validation via Progressive Knockout}

\begin{figure}[t]
    \centering
    \includegraphics[width=\columnwidth]{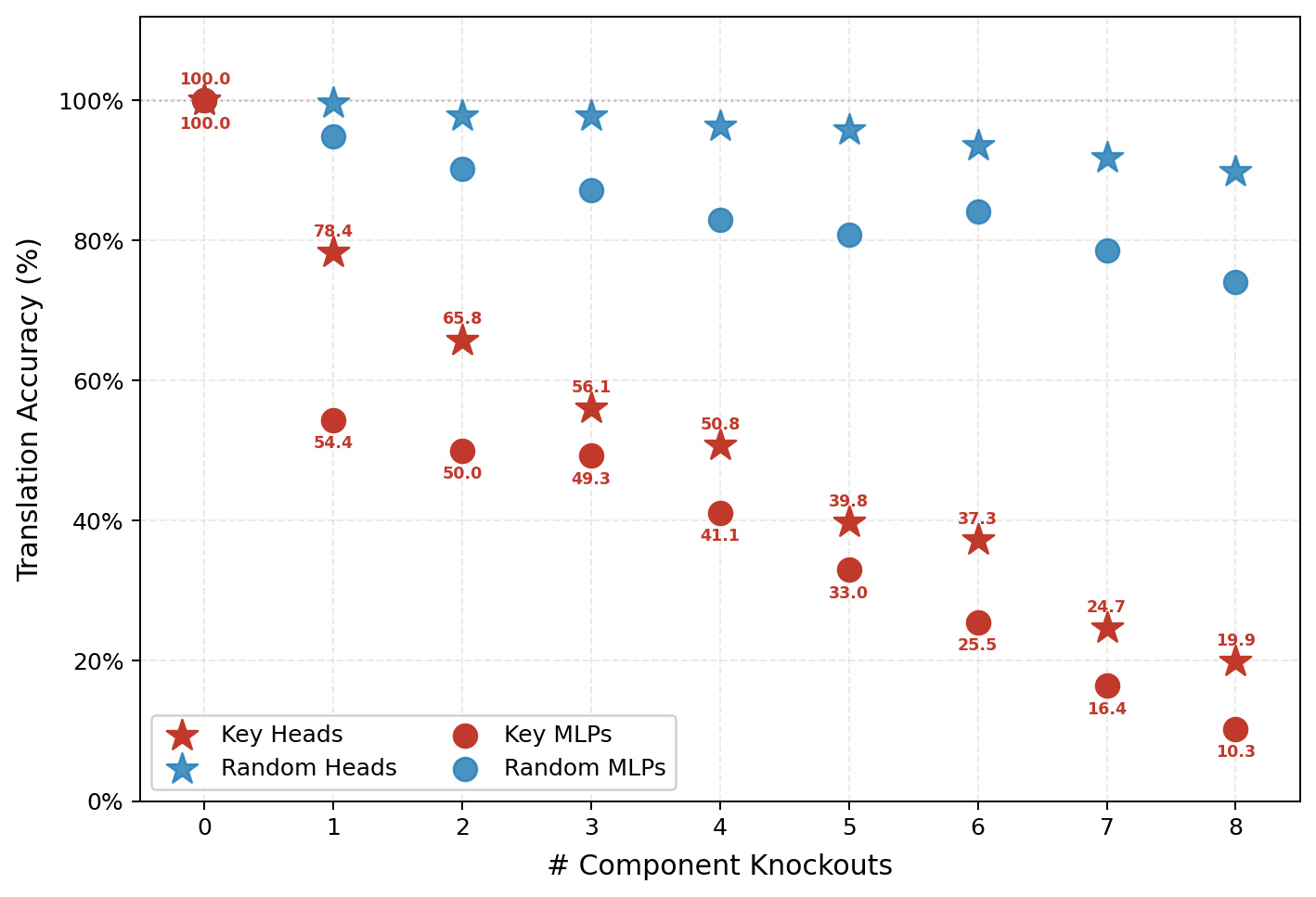}
    \caption{Progressive knockout analysis for Hindi$\rightarrow$Sindhi translation. CIRCE-identified components produce substantially larger degradation than randomly selected components.}
    \label{fig:knockout_hisd}
\end{figure}

We evaluate whether CIRCE-identified components are causally necessary for translation behavior through progressive knockout experiments, where identified attention heads and MLP blocks are sequentially ablated using mean-activation replacement. As shown in Figure~\ref{fig:knockout_hisd}, ablating CIRCE-identified components produces substantially larger degradation than ablating randomly selected components. To rule out circularity from targeted optimization, we additionally evaluate the same CIRCE-identified components in a general LoRA SFT+DPO model trained without restricting optimization to these components. Ablating eight CIRCE-identified components reduces BLEU from 25.53 to 16.54 ($35.2\%$), whereas ablating eight randomly selected components reduces BLEU only from 25.53 to 25.41 ($0.45\%$). For Hindi$\rightarrow$Sindhi, removing a single key MLP reduces translation accuracy from 100\% to 54.4\%, whereas random MLP ablation causes only minor degradation. We further observe that attention-head ablation leads to gradual degradation, indicating distributed routing behavior, while MLP ablation causes sharper collapse, suggesting concentrated semantic transformation functions. Similar trends for Hindi$\rightarrow$Urdu are provided in Appendix Section~\ref{app:causal}.
\newline

\subsection{Attention Routing Behavior}

We next analyze the most causally important attention heads identified by CIRCE to understand how multilingual translation routing is implemented. As shown in Appendix Figures~\ref{fig:appendix_attention_hisd} and~\ref{fig:appendix_attention_hiur}, both translation directions exhibit a consistent routing pipeline consisting of structural anchoring, source-language identification, translation-direction routing, and final target-language commitment.

Early-layer heads primarily attend to formatting boundaries and language identifiers, whereas later layers shift toward translation-direction routing and target-language commitment. Unlike prior English-centric analyses where attention heads predominantly align to source content tokens, the discovered Indic translation circuits rely heavily on explicit language-label anchoring. We attribute this behavior to the strong lexical and subword overlap between closely related Indic languages, where language identifiers provide more discriminative routing signals than raw lexical content. Additional head visualizations and layer-wise routing analyses are provided in Appendix Section ~\ref{app:attention}.

\subsection{MLP Representation Dynamics}

To analyze semantic processing within the translation circuit, we probe MLP hidden states using cosine similarity with source-language, target-language, and translation-indicator token embeddings. As shown in Appendix Figures~\ref{fig:appendix_mlp_hisd} and~\ref{fig:appendix_mlp_hiur}, source-language alignment gradually increases through middle layers, whereas target-language alignment emerges primarily in the final layers, indicating delayed target-language realization during multilingual translation. ; We further observe consistently negative English alignment across all layers for both translation directions, contradicting previously reported English-centric pivot mechanisms. Instead, dominant alignment emerges along shared Indic semantic directions, suggesting that closely related Indic translation operates through a shared semantic subspace rather than an explicit pivot language. Additional layer-wise analyses are provided in Appendix Section~\ref{app:mlp}.

\subsection{Sparse Feature Behavior}

Finally, we analyze Sparse Autoencoder (SAE) features extracted from the residual stream to study how translation-relevant representations are distributed across layers. As shown in Appendix Figures~\ref{fig:appendix_sae_hisd} and~\ref{fig:appendix_sae_hiur}, both translation directions exhibit strongly layer-localized feature distributions, where target-language-specific features dominate early layers while Hindi-specific features emerge more prominently in later layers.

Importantly, highly selective features do not necessarily exhibit strong causal influence. Instead, translation behavior is governed by a sparse subset of high-lift features distributed across both source and target language populations. Together with the attention and MLP analyses, these findings suggest that multilingual translation emerges through sparse routing circuits combined with shared semantic computation rather than through explicit pivot-language translation. Additional feature-level analyses are provided in Appendix Section~\ref{app:sparse}.

\section{Discussion}
\paragraph{\bf Expert Assessment.}
To complement automatic metrics, we conduct human evaluation using native bilingual experts proficient in Hindi--Urdu and Hindi--Sindhi (Devanagari). Experts independently evaluate randomly sampled translations from the SFT and DPO stages of both Gemma-2 baselines and the \textsc{\model{}} pipeline using adequacy and fluency scores on a 1--5 scale. Table~\ref{tab:expert_assessment} reports the percentage of ratings $\geq 4$. Across both language pairs, expert judgements consistently confirm the effectiveness of \textsc{\model{}} over full-parameter baselines. For Hi--Ur, the final \textsc{\model{}} configuration improves adequacy from 36.7\% to 76.7\% and fluency from 53.3\% to 86.7\% relative to the Gemma-2 DPO baseline. Similar gains are already visible at the CIRCE-guided SFT stage, indicating that improvements arise from sparse-circuit optimization rather than DPO alone.

Hi--Sd exhibits the largest qualitative improvements, with \textsc{\model{}} achieving 93.5\% adequacy and 61.3\% fluency compared to 51.3\% and 25.8\% for the Gemma-2 DPO baseline. Experts further observed that \textsc{\model{}} better preserves semantic meaning and produces more morphologically appropriate translations for technical terminology.

Several limitations remain. Sindhi outputs occasionally retain untranslated Hindi loanwords and exhibit minor Unicode inconsistencies around implosive phoneme characters. Performance also degrades on long compound sentences with multiple embedded clauses, where fluency decreases despite largely preserved adequacy.
\begin{table}[t]
\centering
\footnotesize
\setlength{\tabcolsep}{5pt}
\renewcommand{\arraystretch}{1.12}

\begin{tabular}{llcc}

\toprule

\textbf{Direction}
&
\textbf{System}
&
\textbf{Adequacy}
&
\textbf{Fluency}

\\

\midrule

\multicolumn{4}{c}{\textbf{Hindi $\rightarrow$ Urdu}}
\\

\midrule

Gemma-2 (SFT)
& Baseline
& 33.3
& 20.0
\\

Gemma-2 (DPO)
& Baseline
& 36.7
& 53.3
\\

\textsc{CIRCE} 
& Ours
& 56.7
& 56.7
\\

\rowcolor{green!18}

\textsc{Titan}
& Ours
& \textbf{76.7}
& \textbf{86.7}
\\

\midrule

\multicolumn{4}{c}{\textbf{Hindi $\rightarrow$ Sindhi}}
\\

\midrule

Gemma-2 (SFT)
& Baseline
& 45.2
& 12.9
\\

Gemma-2 (DPO)
& Baseline
& 51.3
& 25.8
\\

\textsc{CIRCE}
& Ours
& 64.8
& 39.4
\\

\rowcolor{green!18}

\textsc{Titan} 
& Ours
& \textbf{93.5}
& \textbf{61.3}
\\

\bottomrule

\end{tabular}
\caption{
Human expert evaluation (\% of ratings $\geq 4$ on a 1--5 scale)
for adequacy and fluency across Hindi--Urdu and Hindi--Sindhi
translation tasks. Best results per direction are highlighted in bold.
}

\label{tab:expert_assessment}

\end{table}

\paragraph{\bf Error Analysis.}
Table~\ref{tab:joint_results} reveals a consistent asymmetry in \model{}'s residual errors. Across all evaluation settings, \model{} underperforms the strongest baseline in only two cases, both involving TER for Hindi--Urdu translation directions. Despite higher TER values, \model{} simultaneously achieves stronger BLEU, ChrF2, and COMET scores on the same directions, indicating that the discovered translation circuits prioritize semantic fidelity over minimum-edit lexical reconstruction. For example, on U$\rightarrow$H, \model{} improves BLEU from 25.78 to 27.01 and COMET from 77.37 to 81.28, while producing higher TER relative to the reference translation. Error patterns are strongest for Urdu-as-source directions, whereas the largest gains emerge for shared-script Hindi--Sindhi translation (+45.86\% H$\rightarrow$S and +48.98\% S$\rightarrow$H). This suggests that sparse circuit transfer is more effective under shared orthographic structure and becomes more challenging under Devanagari--Nastaliq script conversion.

\section{Conclusion}
We introduced \model{}, a mechanistic interpretability-guided framework for parameter-efficient multilingual translation that identifies and optimizes sparse translation-crucial circuits instead of full-parameter adaptation. Experiments on Hindi--Urdu and Hindi--Sindhi show substantial improvements over multilingual baselines while updating fewer than 5\% of parameters. Analyses further suggest that low-resource multilingual translation relies on sparse structured circuits within a shared semantic subspace rather than explicit English-centric pivot representations.

\section{Ethics and Limitations}

While \model{} performs strongly across low-resource and typologically diverse settings, broader validation across additional languages, scripts, and domains remains necessary. Although sparse circuit optimization reduces trainable parameters, causal intervention and sparse feature decomposition remain computationally expensive during circuit discovery. Occasional errors include untranslated loanwords, Unicode inconsistencies, and reduced fluency for long compound sentences. The current analysis also focuses on the identified translation circuits rather than the model's complete multilingual representation space, limiting the scope of conclusions about global multilingual organization. \model{} does not guarantee factually or culturally perfect translations, particularly for rare expressions or specialized domains, and may inherit biases from its underlying pretraining and translation data. Our causal findings should therefore be interpreted within the evaluated models and language settings rather than as universal translation circuits. Further evaluation with broader domains, independent annotators, and additional language families would strengthen these conclusions. All datasets are publicly available research corpora, no personally identifiable information was collected, and human evaluation was conducted by bilingual experts. Real-world deployment should include appropriate human oversight and linguistic validation.

\section*{Acknowledgments}
The work is partially supported by a research project COIL-D at IIIT Delhi, funded by MeitY, Govt of India. The authors also acknowledge the support of the Infosys Foundation through Center of AI (CAI) at IIIT-Delhi.



\clearpage
\appendix

\section*{Appendix}
\label{sec:appendix}

\section{Translation Task Templates and Examples}
\label{app:construction}
\subsection{Dataset Construction}

We construct contrastive translation datasets for two directions: Hindi$\rightarrow$Sindhi (Hi$\rightarrow$Sd, Devanagari) and Hindi$\rightarrow$Urdu (Hi$\rightarrow$Ur). The two directions dissociate factors that are typically conflated in multilingual circuit analysis. Hi$\rightarrow$Sd involves no script boundary, thereby isolating cross-lingual semantic mapping from orthographic transduction while preserving a non-trivial transfer challenge through Sindhi-specific implosives and Persian--Arabic lexical borrowing. In contrast, Hi$\rightarrow$Ur crosses a script boundary (Devanagari $\rightarrow$ Perso-Arabic Nastaliq) and, given the strong grammatical overlap between Hindi and Urdu, tests whether the identified circuit encodes semantic content independently of surface orthography.

Contrastive prompts are sampled from the SFT splits. We sample 10K pairs per direction.

\subsection{Prompt Format}

The general template, introduced in Section~4.1 (Eq.~1), is:

\begin{equation}
X^{+} = \texttt{[SRC]: [text] $\rightarrow$ [TGT]:}
\end{equation}

where \texttt{[SRC]} and \texttt{[TGT]} are language-name tokens, and the trailing colon serves as the generation trigger and query position for attention analyses. We use the arrow glyph $\rightarrow$ because it tokenizes as a single token under the Gemma-2 SentencePiece tokenizer. By contrast, the hyphen (\texttt{-}) was empirically verified to fragment at the directive position, destabilizing query-position alignment.

\subsection{Contrastive Pair Construction}

Each source sentence $s$ yields one positive prompt $X^{+}$ and four counterfactual prompts $\{X^{-}_{k}\}_{k=1}^{4}$, differing from $X^{+}$ only in the directive region. Construction follows two principles:

\begin{enumerate}
    \item \textbf{Structural isomorphism}: identical token-position layout, punctuation, and source-sentence span.
    \item \textbf{Targeted lexical substitution}: a single semantic component of the directive is replaced.
\end{enumerate}

Each perturbation type is instantiated with three lexical variants (e.g., Target Nullification: \texttt{Nothing} / \texttt{Null} / \texttt{Void}); reported results are averaged across variants.

Table~\ref{tab:contrastive-taxonomy} illustrates the four perturbation types with representative examples from both translation directions.

\begin{table*}[t]
\centering
\scriptsize
\setlength{\tabcolsep}{4pt}
\renewcommand{\arraystretch}{1.12}

\begin{tabular}{p{2.8cm} p{1.5cm} p{5.2cm} p{3.2cm}}
\toprule
\textbf{Perturbation Type} &
\textbf{Direction} &
\textbf{$X^{+}$ / $X^{-}$ Example} &
\textbf{Ablated Component}
\\
\midrule

\multirow{4}{*}{Target Nullification}

& Hi$\rightarrow$Sd
& \texttt{Hindi: ... $\rightarrow$ Sindhi:}
& \multirow{2}{*}{Target language identity}
\\

&
& \texttt{Hindi: ... $\rightarrow$ Nothing:}
&
\\

& Hi$\rightarrow$Ur
& \texttt{Hindi: ... $\rightarrow$ Urdu:}
& \multirow{2}{*}{Target language identity}
\\

&
& \texttt{Hindi: ... $\rightarrow$ Nothing:}
&
\\

\midrule

\multirow{4}{*}{Action Distortion}

& Hi$\rightarrow$Sd
& \texttt{Hindi: ... $\rightarrow$ Sindhi:}
& \multirow{2}{*}{Source-side instruction frame}
\\

&
& \texttt{Eat the following: ... $\rightarrow$ Sindhi:}
&
\\

& Hi$\rightarrow$Ur
& \texttt{Hindi: ... $\rightarrow$ Urdu:}
& \multirow{2}{*}{Source-side instruction frame}
\\

&
& \texttt{Eat the following: ... $\rightarrow$ Urdu:}
&
\\

\midrule

\multirow{4}{*}{Semantic Obfuscation}

& Hi$\rightarrow$Sd
& \texttt{Hindi: ... $\rightarrow$ Sindhi:}
& \multirow{2}{*}{Translation-task semantics}
\\

&
& \texttt{Hindi: ... $\rightarrow$ Color:}
&
\\

& Hi$\rightarrow$Ur
& \texttt{Hindi: ... $\rightarrow$ Urdu:}
& \multirow{2}{*}{Translation-task semantics}
\\

&
& \texttt{Hindi: ... $\rightarrow$ Color:}
&
\\

\midrule

\multirow{4}{*}{Paradox Insertion}

& Hi$\rightarrow$Sd
& \texttt{Hindi: ... $\rightarrow$ Sindhi:}
& \multirow{2}{*}{Coherent target slot}
\\

&
& \texttt{Hindi: ... $\rightarrow$ Silent Rock:}
&
\\

& Hi$\rightarrow$Ur
& \texttt{Hindi: ... $\rightarrow$ Urdu:}
& \multirow{2}{*}{Coherent target slot}
\\

&
& \texttt{Hindi: ... $\rightarrow$ Silent Rock:}
&
\\

\bottomrule
\end{tabular}

\caption{
Contrastive perturbation taxonomy with representative examples.
Each $X^{-}$ alters a single semantic component of the directive
while preserving structural isomorphism with $X^{+}$.
}
\label{tab:contrastive-taxonomy}

\end{table*}

\subsection{Validation Protocol}

Templates are verified along two axes: \emph{surface fluency} and \emph{task-level disruption}. Because the four perturbations alter the directive at different structural depths, a uniform fluency threshold is inappropriate; accordingly, we apply differentiated validation criteria.

Fluency-preserving perturbations (Semantic Obfuscation and Target Nullification) substitute a single target-slot token and are required to preserve surface fluency. Measured perplexity deviations relative to positive prompts confirm that these perturbations alter directive semantics without disrupting grammatical structure.

Structurally invasive perturbations (Paradox Insertion and Action Distortion) modify the directive at deeper structural levels and are therefore expected to induce higher perplexity by design. For these perturbations, validity is evaluated through task disruption rather than a fluency threshold.

For task-level disruption, all four perturbation types are required to yield zero translation accuracy on a held-out evaluation set, together with strong inter-annotator agreement between two annotators regarding whether each counterfactual successfully removes the translation directive. Templates failing the task-disruption criterion are discarded.

\subsection{Token Taxonomy}

Each token position is assigned one of four functional roles for attention and SAE analyses (Section~4.2).
\newline
IND tokens form the circuit's structural scaffold and draw spike attention from Indicator Heads. SRC tokens are the primary attention targets for Source Heads, while the TGT position serves as the readout site for logit-difference and unembedding analyses. For Hi$\rightarrow$Sd, samples containing Sindhi-specific implosive phonemes are additionally tagged at the SRC level to support phonological circuit analysis.

\begin{table*}[t]
\centering
\small
\setlength{\tabcolsep}{6pt}
\renewcommand{\arraystretch}{1.12}

\begin{tabular}{p{1.8cm} p{4.8cm} p{5.0cm}}
\toprule

\textbf{Role} &
\textbf{Tokens} &
\textbf{Function}
\\

\midrule

\textbf{IND} \\
(Indicator)

&
\texttt{Hindi}, \texttt{Sindhi}, \texttt{Urdu},
$\rightarrow$, \texttt{:}

&
Translation direction and structural scaffold
\\

\textbf{SRC} \\
(Source)

&
Content tokens of the source sentence

&
Semantic payload
\\

\textbf{TGT} \\
(Target)

&
First generated token(s) at query position

&
Translation output signal
\\

\textbf{RAND} \\
(Random)

&
Position-shuffled baseline

&
Control for positional effects
\\

\bottomrule
\end{tabular}

\caption{
Token role definitions used in attention and SAE analyses.
}
\label{tab:token-taxonomy}

\end{table*}

\FloatBarrier

\section{Experimental Setup}
All Baseline fine-tuned models use identical training splits with LoRA rank 16 and batch size 8, with Gemma-2 9B serving as the underlying base model for CIRCE and \model{}. SFT models are trained for 3 epochs, while DPO optimization is performed for 2 epochs on NVIDIA A100 40GB GPUs. \model{} further incorporates Sparse Autoencoder decomposition with latent dimension 14336 for translation-specific feature discovery prior to targeted optimization.

\section{Causal Validation via Progressive Knockout}
\label{app:causal}
\begin{figure}[t]
    \centering
    \includegraphics[width=\columnwidth]{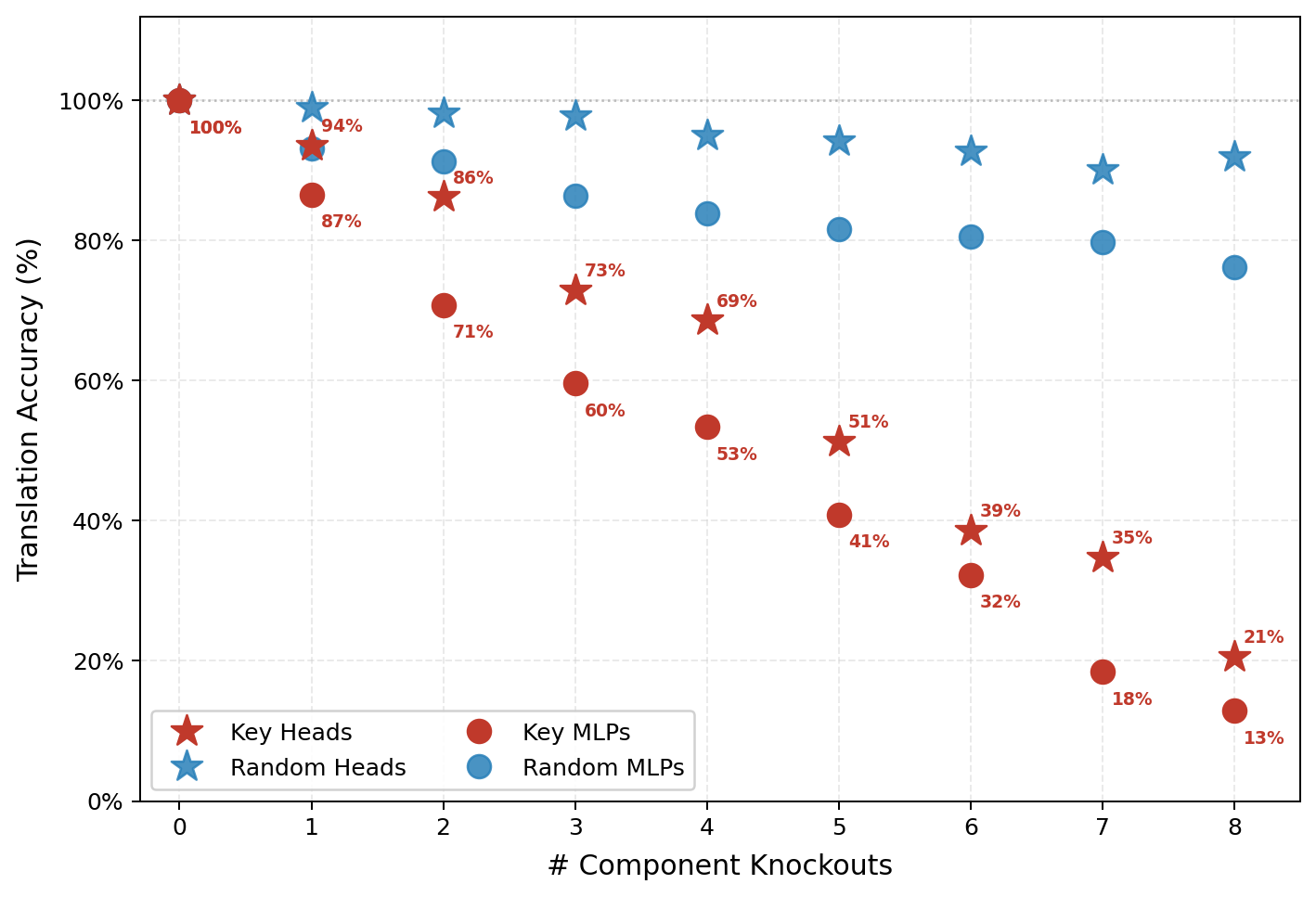}
    \caption{Progressive knockout analysis for Hindi$\rightarrow$Urdu translation. CIRCE-identified components produce substantially larger degradation than randomly selected components.}
    \label{fig:appendix_knockout_hiur}
\end{figure}

In addition to the Hindi$\rightarrow$Sindhi results presented in the main paper, Figure~\ref{fig:appendix_knockout_hiur} presents the corresponding progressive knockout analysis for Hindi$\rightarrow$Urdu translation. Across both translation directions, ablating CIRCE-identified components consistently produces substantially larger degradation than ablating randomly selected attention heads and MLP blocks, confirming that \model{} identifies causally meaningful translation circuits rather than incidentally activated neurons.

For Hindi$\rightarrow$Urdu, removing the first key MLP reduces translation accuracy from 100\% to approximately 87\%, whereas random MLP ablation causes only marginal degradation. Similar to Hindi$\rightarrow$Sindhi, attention-head ablation results in gradual performance decline, indicating distributed routing behavior across multiple heads, while MLP ablation causes sharper early-stage collapse, suggesting concentrated semantic transformation functions.

Across both language pairs, the gap between CIRCE-identified and random components widens monotonically as additional components are removed. Even after multiple interventions, randomly selected components preserve most translation capability, whereas CIRCE-identified circuits rapidly collapse toward near-chance performance. Together, these findings further support that multilingual translation behavior is highly localized within sparse transformer subcircuits.
\section{Attention Routing Behavior}
\label{app:attention}

\begin{figure*}[t]
    \centering
    \includegraphics[width=\textwidth]{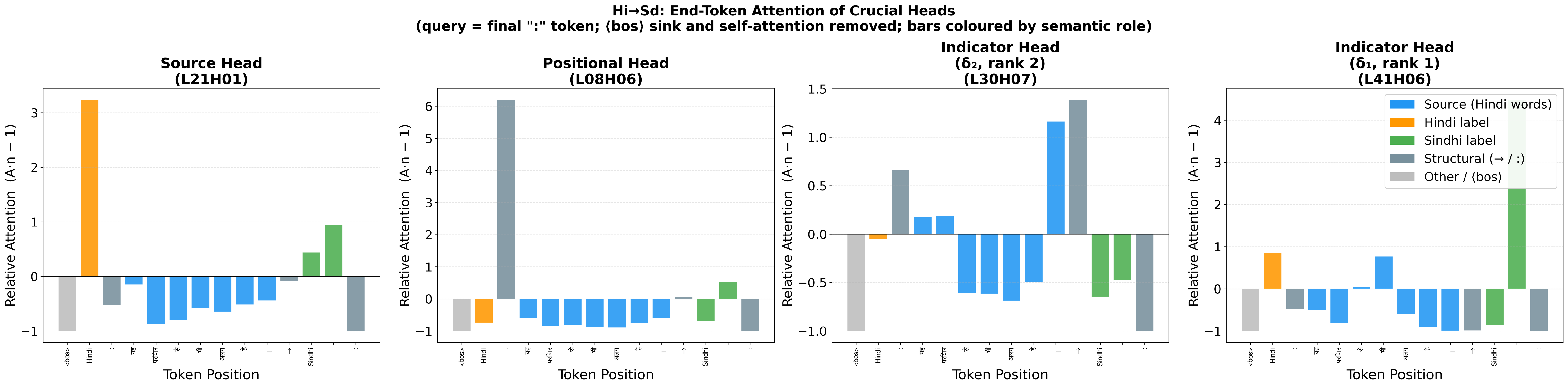}
    \caption{Dominant CIRCE-identified attention heads for Hindi$\rightarrow$Sindhi translation. Different heads specialize in structural anchoring, source-language identification, translation-direction routing, and target-language commitment.}
    \label{fig:appendix_attention_hisd}
\end{figure*}

\begin{figure*}[t]
    \centering
    \includegraphics[width=\textwidth]{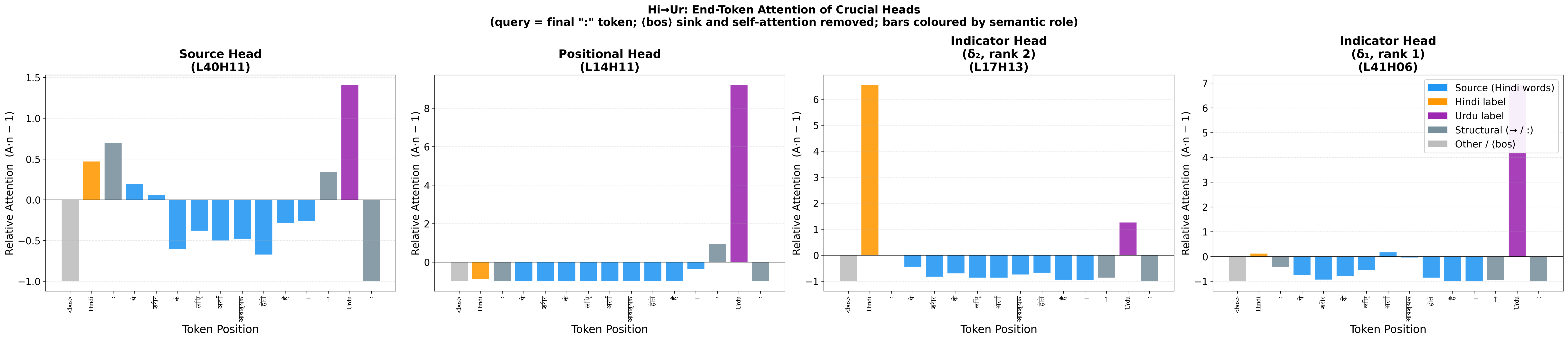}
    \caption{Dominant CIRCE-identified attention heads for Hindi$\rightarrow$Urdu translation. Compared to Hindi$\rightarrow$Sindhi, the Urdu circuit exhibits broader mid-layer routing behavior associated with script-conversion complexity.}
    \label{fig:appendix_attention_hiur}
\end{figure*}

Figures~\ref{fig:appendix_attention_hisd} and~\ref{fig:appendix_attention_hiur} present the dominant CIRCE-identified attention heads for Hindi$\rightarrow$Sindhi and Hindi$\rightarrow$Urdu translation respectively. Across both language pairs, we observe a highly structured routing pipeline in which early-layer heads primarily attend to prompt separators and formatting boundaries, followed by mid-layer heads specializing in source-language identification through explicit language labels. Later layers progressively shift toward translation-direction routing and final target-language commitment.

In both translation directions, the strongest causal heads emerge near the final generation layers, particularly around the terminal translation commitment stage. Notably, several heads consistently attend to language-indicator tokens rather than source content tokens, suggesting that routing decisions are strongly conditioned on explicit language identifiers. This behavior differs substantially from previously reported English-centric translation circuits, where attention heads predominantly align to source semantic content.

Although both translation directions exhibit similar routing structure, the Hindi$\rightarrow$Urdu circuit displays broader and more distributed mid-layer routing behavior compared to Hindi$\rightarrow$Sindhi. We attribute this difference to the additional script-conversion complexity introduced by Urdu Nastaliq generation. Overall, these findings suggest that low-resource Indic translation relies heavily on label-aware routing mechanisms operating at prompt boundaries before semantic transformation occurs within downstream MLP representations.

\section{MLP Representation Dynamics}
\label{app:mlp}

\begin{figure*}[t]
    \centering
    \includegraphics[width=\textwidth]{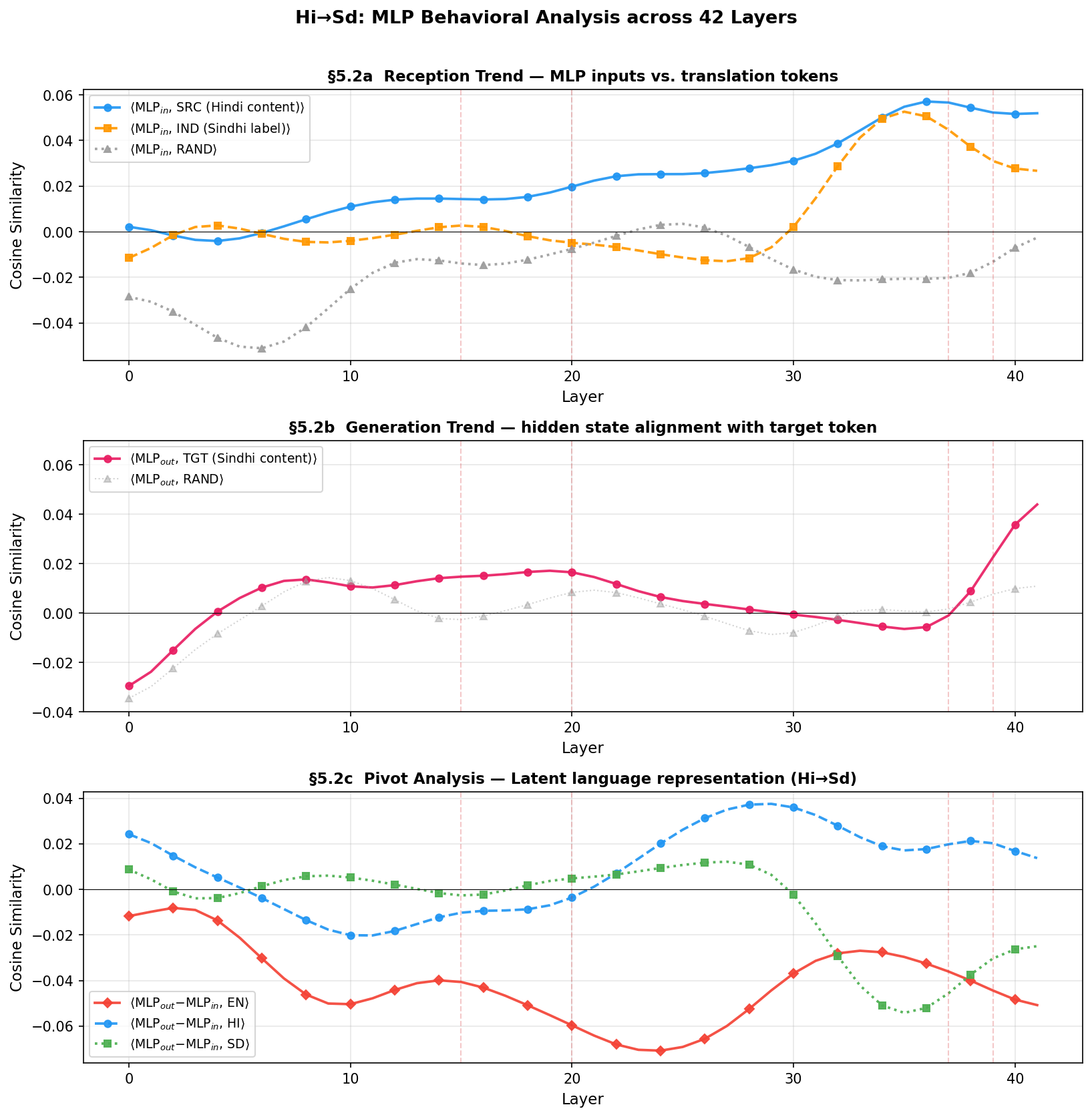}
    \caption{MLP representation dynamics for Hindi$\rightarrow$Sindhi translation. Top: source-language reception behavior. Middle: target-language generation behavior. Bottom: latent pivot analysis across Hindi, Sindhi, and English representation directions.}
    \label{fig:appendix_mlp_hisd}
\end{figure*}

\begin{figure*}[t]
    \centering
    \includegraphics[width=\textwidth]{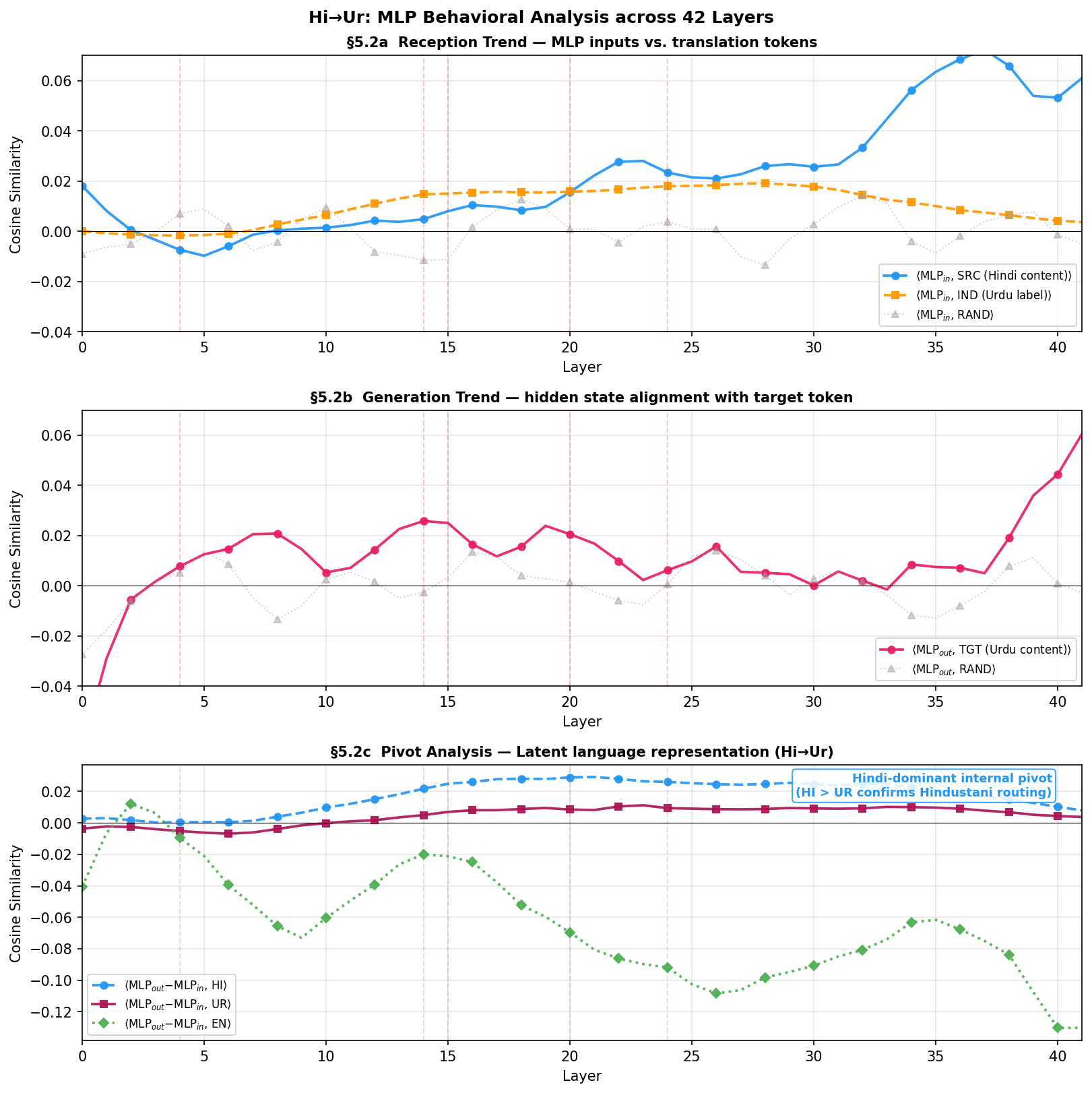}
    \caption{MLP representation dynamics for Hindi$\rightarrow$Urdu translation. Similar to Hindi$\rightarrow$Sindhi, translation behavior evolves through shared Indic semantic directions rather than explicit English pivots.}
    \label{fig:appendix_mlp_hiur}
\end{figure*}

Figures~\ref{fig:appendix_mlp_hisd} and~\ref{fig:appendix_mlp_hiur} present the layer-wise MLP representation dynamics for Hindi$\rightarrow$Sindhi and Hindi$\rightarrow$Urdu translation respectively. To analyze semantic processing within the translation circuit, we project MLP hidden states onto source-language, target-language, and translation-indicator token embeddings using cosine similarity.

Across both translation directions, source-language alignment gradually increases through middle layers and peaks near the later network stages, indicating progressive encoding of source-language semantics within MLP representations. In contrast, target-language alignment remains weak for most layers before sharply increasing near the final generation layers, suggesting that multilingual translation postpones target-language realization until late-stage decoding.

We further analyze whether translation proceeds through an implicit English pivot, as suggested in prior multilingual interpretability studies. Projecting MLP transformation vectors onto English, Hindi, Urdu, and Sindhi representation directions reveals consistently negative English alignment across nearly all layers for both language pairs. This strongly contradicts previously reported English-centric multilingual pivot mechanisms.

Instead, the dominant latent alignment emerges along shared Indic semantic directions, particularly Hindi-centered representations, suggesting that closely related Indic translation directions operate through a shared Hindustani semantic subspace rather than through any explicit pivot language. In this shared latent space, semantic computation becomes approximately orthogonal to explicit language identifiers, enabling language-neutral processing prior to final target-language commitment.

\section{Sparse Feature Behavior}
\label{app:sparse}

\begin{figure*}[t]
    \centering
    \includegraphics[width=\textwidth]{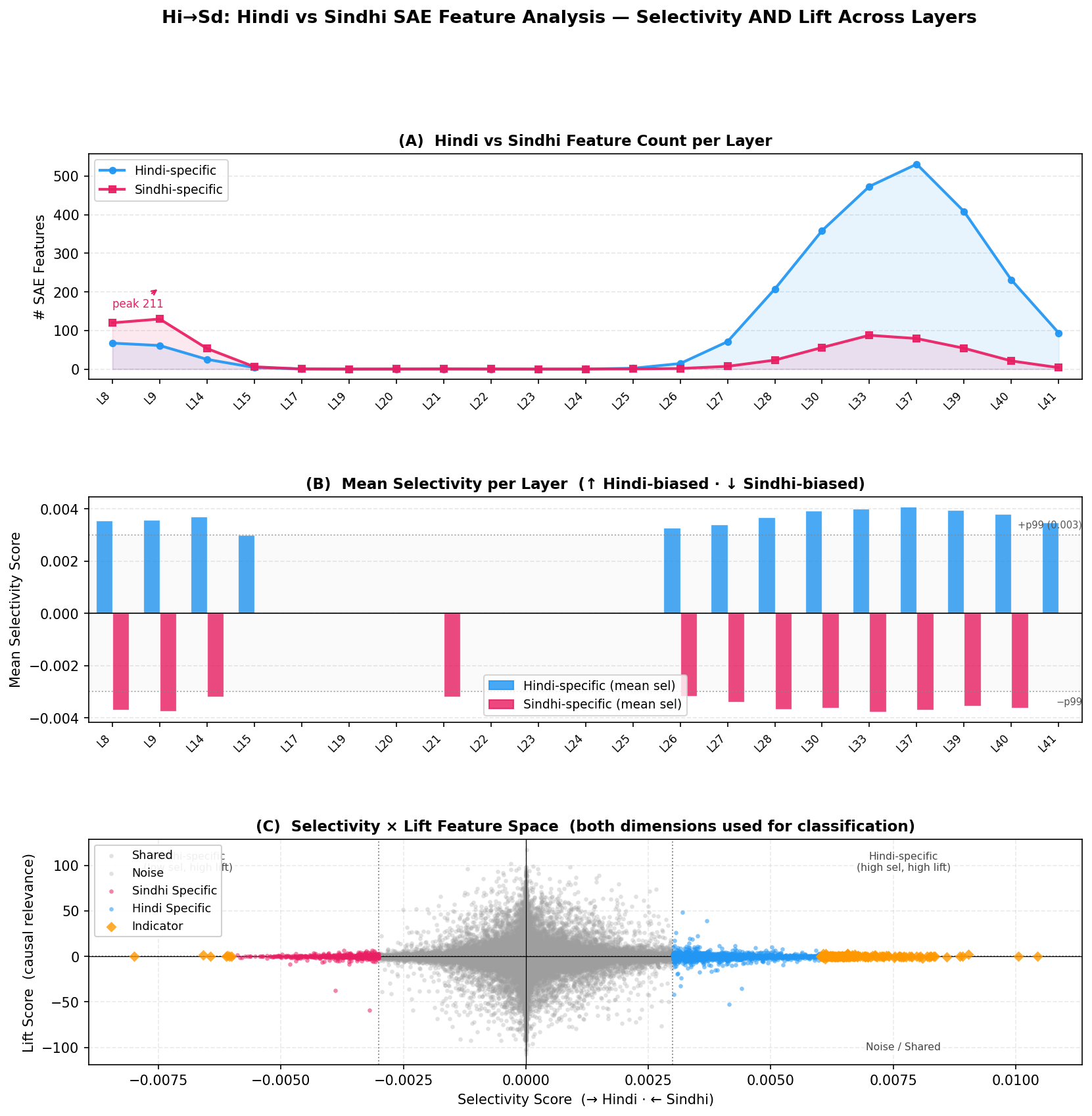}
    \caption{Sparse feature analysis for Hindi$\rightarrow$Sindhi translation. Top: layer-wise distribution of Hindi-specific and Sindhi-specific SAE features. Middle: mean feature selectivity across layers. Bottom: joint selectivity--lift feature space highlighting sparse causally relevant translation features.}
    \label{fig:appendix_sae_hisd}
\end{figure*}

\begin{figure*}[t]
    \centering
    \includegraphics[width=\textwidth]{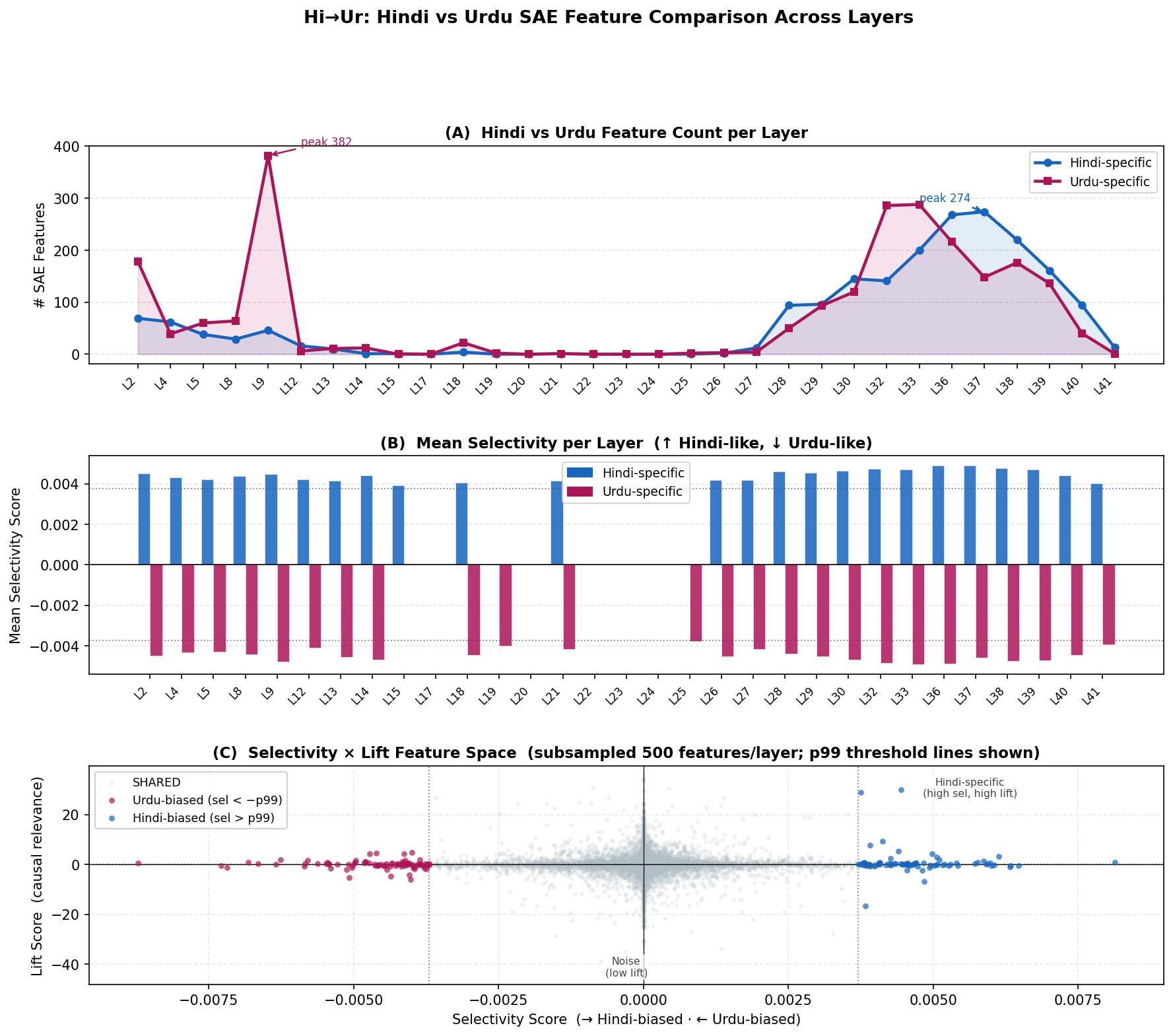}
    \caption{Sparse feature analysis for Hindi$\rightarrow$Urdu translation. Similar to Hindi$\rightarrow$Sindhi, causally important translation behavior is concentrated within a sparse subset of high-lift features.}
    \label{fig:appendix_sae_hiur}
\end{figure*}

Figures~\ref{fig:appendix_sae_hisd} and~\ref{fig:appendix_sae_hiur} present the layer-wise Sparse Autoencoder (SAE) feature distributions for Hindi$\rightarrow$Sindhi and Hindi$\rightarrow$Urdu translation respectively. To analyze translation-relevant representations, we classify SAE features jointly using selectivity and causal-lift statistics in order to distinguish routing-oriented features from causally important translation features.

Across both language pairs, we observe strongly layer-localized feature distributions in which target-language-specific features dominate earlier layers while Hindi-specific features emerge more prominently in later layers. Between these regions lies a middle-layer transition zone where neither language exhibits strong dominance, consistent with the shared latent-space behavior observed in the MLP analysis.

Importantly, highly selective features do not necessarily exhibit strong causal influence. Most language-specific features produce minimal causal lift despite strong selectivity scores, indicating that activation specialization alone is insufficient to explain translation behavior. Instead, translation is governed by a sparse subset of high-lift features distributed across both source and target language populations.

Combining the SAE findings with the attention and MLP analyses yields a unified mechanistic interpretation of low-resource Indic translation. Attention heads primarily implement language-aware routing and prompt-boundary coordination, whereas MLP layers perform semantic transformation within a shared Hindustani latent subspace. Translation behavior therefore emerges through sparse routing circuits coupled with language-neutral semantic computation rather than through explicit pivot-language translation.

\end{document}